\pdfoutput=1

\documentclass[11pt]{article}

\usepackage[]{acl}

\usepackage{times}
\usepackage{latexsym}

\usepackage[T1]{fontenc}

\usepackage[utf8]{inputenc}

\usepackage{microtype}

\usepackage{graphicx}
\graphicspath{{Figure/}}
\usepackage{amsmath}
\usepackage{amssymb}
\usepackage{amsfonts}
\usepackage{enumitem}
\usepackage{booktabs}
\usepackage{mathrsfs}
\usepackage{multirow}
\usepackage{arydshln}
\usepackage{bm}
\usepackage{algorithm}
\usepackage{subfigure}
\usepackage{multicol}
\usepackage[noend]{algpseudocode}

\title{Deep RL with Hierarchical Action Exploration for Dialogue Generation}

\author{Itsugun Cho$^1$ \quad Ryota Takahashi$^1$ \quad Yusaku Yanase$^1$ \quad Hiroaki Saito$^1$ \\[3pt]
Keio University, Japan$^1$ \\
{\tt \{choitsugun, ryota.0226.tokky, y.y32851107\}@keio.jp}\\ \tt hxs@ics.keio.ac.jp\\}

\begin{document}
\maketitle
\begin{abstract}
Traditionally, approximate dynamic programming is employed in dialogue generation with greedy policy improvement through action sampling, as the natural language action space is vast. However, this practice is inefficient for reinforcement learning (RL) due to the sparsity of eligible responses with high action values, which leads to weak improvement sustained by random sampling. This paper presents theoretical analysis and experiments that reveal the performance of the dialogue policy is positively correlated with the sampling size. To overcome this limitation, we introduce a novel dual-granularity Q-function that explores the most promising response category to intervene in the sampling process. Our approach extracts actions based on a grained hierarchy, thereby achieving the optimum with fewer policy iterations. Additionally, we use offline RL and learn from multiple reward functions designed to capture emotional nuances in human interactions. Empirical studies demonstrate that our algorithm outperforms baselines across automatic metrics and human evaluations. Further testing reveals that our algorithm exhibits both explainability and controllability and generates responses with higher expected rewards.\let\thefootnote\relax\footnotetext{Submission history: [v1] Wed, 22 Mar 2023.}
\end{abstract}

\section{Introduction}
To ensure a satisfactory user experience, an intelligent dialogue agent is required to respond fluently and naturally while being endowed with a “forward-looking” capacity in the dialogue. A predominant approach to training agents is to optimize the maximum likelihood estimation (MLE) objective for the probability distribution of responses. However, this supervised technique is insufficient to learn a long-term behavior since the corpus often contains suboptimal dialogues, and MLE cannot model the future direction of the conversation. Instead, if we view the open-domain dialogue as a control problem, RL could enable agents to automatically adjust policy concerning the pre-defined appraisal functions via a trial-and-error process.

Recent work in RL for dialogue generation is well summarized by \citet{lone2022self}. Most prior studies are built on the actor-critic framework to train a dialogue agent, which optimizes the policy with the support of $N$ generated possible actions from the agent to maximize the action-value function (i.e., Q-function). Although this self-behavior cloning does not rely on policy gradients, thereby avoiding the issue of diverging from human language, it suffers from slow improvement and often falls into a trivial local optimum. We argue that the actions that can make the Q-function produce a higher value (i.e., Q-value) in a given state will be similar at an elevated abstraction rank. For example, in a conversation about favorite food, responses about steak are more likely to get higher cumulative expected rewards than those about business. If we apprehend which abstract category of actions can obtain a higher Q-value, then generating responses of that category for the greedy policy will make training more efficient. To this end, we propose a dual-granularity Q-function to evaluate the utility carried by an action under different levels of abstraction. Specifically, our algorithm contains a coarse-grained Q-function by category-represented responses that aim to lock the optimal category and a fine-grained Q-function by token-represented responses that strive to extract the optimal action. In this manner, the infinite action space is divided into several blocks at the high-level abstraction, enabling exploration of the entire action space to adapt policy on the fly. Since RL requires numerous costly interactions with the environment (i.e., real users), we applied offline RL to our algorithm, which leverages the previously collected dataset DailyDialog \cite{li2017dailydialog} for policy training. Moreover, inspired by the psychology of human conversation, four reward functions are devised to improve the agent’s ability to engage in natural dialogue. Experimental results demonstrate that by training with our algorithm, four state-of-the-art dialogue models achieve a significant performance improvement. The controllability and effectiveness of our approach are clarified in the discussion. Our main contributions in this study are two-fold:\\
(1) To the best of our knowledge, this is the first attempt to implement offline RL by different-grained representations of natural language, which provides a unified algorithmic template for large action space tasks, extending beyond dialogue generation.\\
(2) The quantitative and qualitative empirical verifications have established that our approach exhibits a high level of trustworthiness.

\section{Methodology}
\subsection{Preliminaries}
We begin with a Markov decision process represented by a tuple $M=(S,A,T,R,\gamma)$, where $S$ is the state space, $A$ is the action space, $T$ is the state transition function, $R$ is the reward function, and $\gamma\in(0,1)$ is a discount factor. In the dialogue setting, the agent observes a context $s$, executes its policy $\pi$, by generating a response $a$, according to $\pi(a|s)$, transitions to a new context $s^{\prime}$, and receives a reward $r=R(s,a)$. The goal is to learn $\pi$ to maximize cumulative reward from a dataset $\mathcal{D}$, which consists of multiple $(s,a,r,s^{\prime})$ pairs produced under a potential behavior $\pi_\beta$. Therefore, prior works typically rely on the actor-critic style that alternates between fitting Q-function by the policy evaluation based on approximate dynamic programming (i.e., iterating the Bellman operator via minimizing the temporal difference error) and improving $\pi$ by updating it toward responses that maximize the expected Q-value.
\begin{equation}
\begin{split}
&{\rm Evaluation}:Q\leftarrow\mathop{\rm{arg\,min}}\limits_{Q}\mathbb{E}_{(s,a,r,s^{\prime})\backsim \mathcal{D}}\\
&[(r+\gamma\mathbb{E}_{a^{\prime}\backsim\pi(a^{\prime}|s^{\prime})}[Q(s^{\prime},a^{\prime})]-Q(s,a))^2]
\end{split}
\end{equation}
\begin{equation}
\begin{split}
&{\rm Improvement}:\pi\leftarrow\mathop{\rm{arg\,max}}\limits_{\pi}\mathbb{E}_{s\backsim\mathcal{D},a\backsim\pi(a|s)}\\
&\qquad\qquad\qquad\qquad[Q(s,a)] 
\end{split}
\label{2}
\end{equation}

A challenge in implementing offline RL is that static $\mathcal{D}$ has limited coverage of $S$ and $A$, whereby $\pi$ may be biased toward out-of-distribution (OOD) actions for $\pi_\beta$ with erroneously high Q-value \cite{fujimoto2019off,kumar2020conservative,kostrikovoffline}. Hence, we follow \citet{jang2022gpt} to employ the one-step algorithm \cite{brandfonbrener2021offline} for on-policy evaluation, as shown below, which can iterate in a rather stable manner since actions are always in $\mathcal{D}$ to avoid the OOD due to distribution deviations between $\pi$ and $\pi_\beta$.
\begin{equation}
\begin{split}
&{\rm{Evaluation}}:Q\leftarrow\mathop{\rm{arg\,min}}\limits_{Q}\mathbb{E}_{(s,a,r,s^{\prime},a^{\prime})\backsim \mathcal{D}}\\
&\qquad\quad[(r+\gamma Q(s^{\prime},a^{\prime})-Q(s,a))^2]
\end{split}
\label{3}
\end{equation}

\subsection{Dual-granularity Q-function}
Traditionally, to implement the $arg\,max$ operator in Eq.(\ref{2}), a set of responses is sampled by $\pi(a|s)$, and $\pi$ is updated based on the one that yields the highest action value according to the Q-function. We can show that the renewed policy by more responses has a higher state value (i.e., a better performance). When the sampling size is large enough to cover the entire action space, $\pi$ can theoretically iterate to the optimum. We also show that the renewed policy by responses with a higher Q-value has a higher state value. We formalize the results in Theorem 1 and 2, respectively. The detailed proofs are presented in Appendix \ref{app_A}.\\
\textbf{Theorem 1}. Given a policy $\pi$ and the number of sampled actions $L$, if we update the new policy by
\begin{equation}
\forall s, \,\pi^{\prime}_L=\mathop{\rm{arg\,max}}\limits_{a\in \{a_i\}^L_{i=1}\backsim\pi(a|s)}Q^{\pi}(s,a)
\nonumber
\end{equation}
then for any $N$, $M$, such that $N\ge M\ge 1$, $\forall s$, $V^{\pi^{\prime}_N}(s)\ge V^{\pi^{\prime}_M}(s)$ always holds.\\
\textbf{Theorem 2}. Given the policy $\pi_\alpha$, $\pi_\beta$, and $\pi$, s.t. $\mathbb{E}_{a\backsim\pi_\alpha(a|s)}[Q^{\pi}(s,a)]\ge\mathbb{E}_{a\backsim\pi_\beta(a|s)}[Q^{\pi}(s,a)]$, if the number of sampled actions is $L$, and we update the new policy by
\begin{equation}
\begin{split}
&\forall s, \,\pi^{\prime}_1=\mathop{\rm{arg\,max}}\limits_{a\in \{a_i\}^L_{i=1}\backsim\pi_\alpha(a|s)}Q^{\pi}(s,a)\\
&\forall s, \,\pi^{\prime}_2=\mathop{\rm{arg\,max}}\limits_{a\in \{a_i\}^L_{i=1}\backsim\pi_\beta(a|s)}Q^{\pi}(s,a)
\end{split}
\nonumber
\end{equation}
then $\forall s$, $V^{\pi^{\prime}_1}(s)\ge V^{\pi^{\prime}_2}(s)$ always holds.
\begin{algorithm*}[t]
\renewcommand{\algorithmicrequire}{\textbf{Input:}}
\renewcommand{\algorithmicensure}{\textbf{Output:}}
\caption{Dual-granularity Q-function}\label{algorithm}
\begin{algorithmic}[t]
\Require
\\
The dataset $\mathcal{D}=\{\mathcal{D}_i=(s,a,r,s^{\prime},a^{\prime})\}^M_{i=1}$, the classifier with action category set $\{\bar{a}_i\}^N_{i=1}$.
\Ensure
\\
The agent with policy $\pi_\mu$.
\end{algorithmic} 
\begin{algorithmic}[1]
\State {\textbf{Initialization:}}
\State{\indent Build the dataset $\mathcal{D}^c=\{\mathcal{D}_i^c=(s,\bar{a},r,s^{\prime},\bar{a}^{\prime})\}^M_{i=1}$ base on the dataset $\mathcal{D}$ using the classifier.
\State\indent Initialize the critic and target network parameters $\phi$, $\hat{\phi}$, $\theta$, $\hat{\theta}$, control generator $\pi_\psi$, and agent $\pi_\mu$.
\State\indent Fine-tuning the control generator $\pi_\psi$ using $(s,a,\bar{a})$ triples, where $s$ and $a$ are from dataset $\mathcal{D}$.}
\State{\textbf{for} $i=1$ \textbf{to} until $Q_\phi$ and $Q_\theta$ converge \textbf{do}}
\State{\indent $\#$ The iteration stops for the first converged Q-function, while the rest continue until convergence.
\State\indent $\#$ Policy Evaluation on the dual-granularity Q-function.}
\State{\indent $\phi\leftarrow\mathop{\rm{arg\,min}}\limits_{\phi}(r+\gamma Q_{\hat{\phi}}(s^{\prime},\bar{a}^{\prime})-Q_\phi(s,\bar{a}))^2\quad (s,\bar{a},r,s^{\prime},\bar{a}^{\prime})=\mathcal{D}^c_i$
\State\indent $\theta\leftarrow\mathop{\rm{arg\,min}}\limits_{\theta}(r+\gamma Q_{\hat{\theta}}(s^{\prime},a^{\prime})-Q_\theta(s,a))^2\quad (s,a,r,s^{\prime},a^{\prime})=\mathcal{D}_i$
\State\indent Every $n$ step $\hat{\phi}\leftarrow \phi$, $\hat{\theta}\leftarrow \theta$.}
\State {\textbf{end for}}
\State{\textbf{for} $i=1,s\in\mathcal{D}_i$ \textbf{to} until $\pi_\mu$ converge \textbf{do}}
\State{\indent $\#$ Policy Improvement for the agent.
\vspace{-0.9em}\begin{multicols}{2}
\State\indent $\bar{a}^\ast=\mathop{\rm{arg\,max}}\limits_{\bar{a}}Q_{\phi}(s,\bar{a})\quad\bar{a}\in\{\bar{a}_i\}^{N}_{i=1}\quad$
\State\indent $\{a_i\}^L_{i=1}=\pi_\psi(a|s,\bar{a}^\ast)$
\State\indent $a^\ast=\mathop{\rm{arg\,max}}\limits_{a}Q_{\theta}(s,a)\quad a\in\{a_i\}^L_{i=1}$
\State\indent $\mu\leftarrow\mathop{\rm{arg\,min}}\limits_{\mu}-{\rm log}\,\pi_\mu(a^\ast|s)$\end{multicols}}
\vspace{-0.9em}\State {\textbf{end for}}
\end{algorithmic}
\label{a1}
\end{algorithm*}

Since it is impractical to exhaust all possible responses, we focus on constructing the sampling process in a more organized manner instead of randomly in order to yield responses with a higher Q-value and learn an agent with better performance with the same sample size. We call response $a$ the fine-grained action and its category representation $\bar{a}$ the coarse-grained action, where $\bar{a}$ belongs to a finite set of categories $\{\bar{a}_i\}^N_{i=1}$ and $\bar{a}$ can obtain by ${\rm arg\,max}_{\bar{a}_i}F(\bar{a}_i|a)$, where $F$ is a classifier. The coarse-grained Q-function searches the category $\bar{a}^\ast$ with the highest Q-value from $\{\bar{a}_i\}^N_{i=1}$, where policy evaluation is as follows.
\begin{equation}
\begin{split}
&\phi\leftarrow\mathop{\rm{arg\,min}}\limits_{\phi}\mathbb{E}_{(s,\bar{a},r,s^{\prime},\bar{a}^{\prime})\backsim\mathcal{D}^c}\\
&[(r+\gamma Q_{\hat{\phi}}(s^{\prime},\bar{a}^{\prime})-Q_\phi(s,\bar{a}))^2]
\end{split}
\end{equation}
where $\phi$ and $\hat{\phi}$ are the parameters of the critic and target networks \cite{mnih2015human}, and the same is true for $\theta$ and $\hat{\theta}$ in Eq.(\ref{5}). $\mathcal{D}^c$ is a new dataset built on $\mathcal{D}$ that replaces fine-grained actions with coarse-grained actions. Then, a fine-tuned control generator with policy $\pi_\psi$ generates a set of responses $\{a_i\}^L_{i=1}$ under the specified category according to $\pi_\psi(a|s,\bar{a}^\ast)$. The fine-grained Q-function selects the response $a^\ast$ with the highest Q-value from $\{a_i\}^L_{i=1}$, where policy evaluation is as follows.
\begin{equation}
\begin{split}
&\theta\leftarrow\mathop{\rm{arg\,min}}\limits_{\theta}\mathbb{E}_{(s,a,r,s^{\prime},a^{\prime})\backsim\mathcal{D}}\\
&[(r+\gamma Q_{\hat{\theta}}(s^{\prime},a^{\prime})-Q_\theta(s,a))^2]
\end{split}
\label{5}
\end{equation}
Finally, the agent with policy $\pi_\mu$ is optimized by
\begin{equation}
\mu\leftarrow\mathop{\rm{arg\,min}}\limits_{\mu}\mathbb{E}_{s\backsim\mathcal{D}}[-{\rm log}\,\pi_\mu(a^\ast|s)]
\end{equation}
Our pseudocode is presented in Algorithm \ref{a1}.

\subsection{Rewards}
Our goal is to develop an agent that is sufficiently intrinsically motivated to enrich the interactive content by capturing affective cues in human reactions. To achieve this, we devised four reward functions that assess how empathetic the response is to the conversation. (1) The average cosine similarity between the agent’s response and dull responses. An expression that lacks emotional engagement may limit the development of dialogue. (2) The outpouring of the emotion of surprise. It benefits to build trust and hold the partner’s attention throughout the conversation \cite{shum2018eliza}. (3) The length of response (i.e., the number of tokens in a response). It is a critical signal of engagement in conversation \cite{zhou2020design}. (4) Asking questions. It is an active listening skill that links to conversation management and responsiveness \cite{bodie2012listening}. The total reward was used as $r$, and more details on the scoring design are presented in Appendix \ref{app_B}.

\section{Experiments}
\subsection{Corpus}
We evaluated our approach on the DailyDialog dataset, which was crawled from websites that serve English dialogue in daily life. This dataset is human-rewritten and manually labeled with communication intention and emotion. We referred to its labels of action and emotion for assigning rewards (2) and (4) designed in Section 2.3 to each response. This dataset contains 11,118 / 1,000 / 1,000 multi-turn dialogues for train / test / dev, respectively. We used the set of train and dev for Q-function training and fine-tuning agents and the set of test for evaluation and discussion.

\subsection{Agents}
The following four state-of-the-art generative methods were considered as agents in our experiments. \textbf{GPT-2} is an unsupervised autoregressive language model for textual generation proposed by \citet{radford2019language}. \textbf{DialoGPT} is a pre-trained dialogue model proposed by \citet{zhang2020dialogpt}. This model is based on GPT-2, using the Reddit comments dataset. \textbf{T5} is a unified framework proposed by \citet{raffel2020exploring} that converts all text-based language tasks into a text-to-text format via the transfer learning technique. \textbf{GODEL} is a pre-trained dialogue model proposed by \citet{peng2022godel}. This model is based on T5, using the Reddit discussion dataset. All the agents used the base version of the corresponding pre-trained model.

\subsection{Implementation}
The critic and target networks are the BERT models \cite{kenton2019bert} with a fully connected head on top. The classifier\footnote{It is available in the official repository of Cardiff NLP\,:\\\url{https://huggingface.co/cardiffnlp}\\
Note that many other ways for the category decision are also feasible, such as clustering, LDA, etc. Since TweetTopic and DailyDialog were both crawled from social networking, we consider this classifier is more appropriate for our test.} is a RoBERTa model fine-tuned on the TweetTopic dataset \cite{dimosthenis-etal-2022-twitter}, which divides the responses into 19 pre-defined topics as its action category. Further, the control generator is initialized by the corresponding agent model. To drive the control generator to respond for the specified category, we append the category representation at the beginning of the input for GPT-2 and DialoGPT during the learning and inference, the injection scheme for which followed \citet{cho-etal-2022-personalized}. In addition, for T5 and GODEL, we added the category representation into the task prefix of the T5 framework during the learning and inference. The task prefix was set as “Instruction: given a dialog context, you need to respond related to <category>.” Our implementation was based on PyTorch \cite{paszke2019pytorch} and HuggingFace libraries \cite{wolf2019huggingface}. 

All agents and the control generator were fine-tuned before executing RL. Thus, early stopping was used to determine the number of training iterations of the best checkpoint, and the patience time was set as 5. The batch size was fixed at 32. The Adam algorithm \cite{kingma2015adam} was utilized for optimization, with a learning rate of 2.6e-5 and a warmup step of 6000. The control generator constructs the actions using multinomial sampling with a temperature setting of 1.5 to collect diverse responses. In addition, the update rate of the target network is set as 2.4e-5. The synchronized interval for the target networks was 30 steps. The discount factor was set as 0.5. We considered the network to have converged and terminated the iteration when the change in the loss for 10 consecutive epochs of the target network is less than 0.01.

\subsection{Evaluation}
\begin{table*}[t]
\centering
\begin{tabular}{cccccccccc}
\toprule
\multirow{2}[2]{*}{\textbf {Agent}}&
\multirow{2}[2]{*}{\textbf {Training Method}}&
\multicolumn{4}{c}{\textbf {Dataset-based}}&
\multicolumn{4}{c}{\textbf {Simulator-based}}\cr
\cmidrule(lr){3-6}\cmidrule(lr){7-10}
&&CS $\downarrow$&SE&LR&AQ&CS $\downarrow$&SE&LR&AQ\cr
\toprule
\multirow{3}[2]{*}{\textbf {GPT-2}}
&MLE      &0.712 &0.082 &10.396 &0.308 &0.685 &0.146 &11.276 &0.390\cr
&Standard &0.645 &0.126 &13.020 &0.550 &0.644 &0.206 &13.778 &0.526\cr
&Ours     &\textbf{0.596} &\textbf{0.191} &\textbf{14.463} &\textbf{0.555} &\textbf{0.597} &\textbf{0.238} &\textbf{15.636} &\textbf{0.566}\cr
\midrule
\multirow{3}[2]{*}{\textbf {DialoGPT}}
&MLE      &0.714 &0.069 &9.761  &0.345 &0.687 &0.142 &10.838 &0.492\cr
&Standard &0.645 &0.142 &12.182 &0.579 &0.654 &0.206 &13.772 &0.538\cr
&Ours     &\textbf{0.598} &\textbf{0.171} &\textbf{13.055} &\textbf{0.586} &\textbf{0.588} &\textbf{0.240} &\textbf{14.466} &\textbf{0.604}\cr
\midrule
\multirow{3}[2]{*}{\textbf {T5}}
&MLE      &0.720 &0.063  &9.704 &0.316 &0.651 &0.088 &10.242 &0.396\cr
&Standard &0.621 &0.147 &13.291 &0.532 &0.605 &0.224 &13.676 &0.510\cr
&Ours     &\textbf{0.567} &\textbf{0.202} &\textbf{14.834} &\textbf{0.565} &\textbf{0.553} &\textbf{0.268} &\textbf{15.134} &\textbf{0.552}\cr
\midrule
\multirow{3}[2]{*}{\textbf {GODEL}}
&MLE      &0.718 &0.064  &9.507 &0.318 &0.689 &0.112 &10.132 &0.414\cr
&Standard &0.625 &0.165 &13.553 &0.529 &0.615 &0.235 &13.108 &0.614\cr
&Ours     &\textbf{0.571} &\textbf{0.232} &\textbf{15.272} &\textbf{0.557} &\textbf{0.571} &\textbf{0.258} &\textbf{14.608} &\textbf{0.628}\cr
\bottomrule
\end{tabular}
\caption{\label{t1}For the standard offline RL algorithm and our approach, we use $L=5$ for the number of candidate responses $\{a_i\}^L_{i=1}$. For the simulator-based evaluation, we conducted 1000 dialogues of 5 consecutive turns between the simulator and each method. Each metric is measured per response, and the best score in each metric is in bold. The statistical test revealed that the differences are significant, with a p-value < 0.05.}
\end{table*}
\begin{table*}[t]
\centering
\begin{tabular}{cccccc}
\toprule
\textbf{Agent}& \textbf{Training Method}& \textbf{Quality}& \textbf{Informativeness}& \textbf{Empathy}& \textbf{Engagingness}\\
\toprule
\multirow{3}[2]{*}{\textbf {GPT-2}}
&MLE      &1.4 &1.3 &1.2 &1.1\cr
&Standard &\textbf{1.7} &1.3 &1.4 &1.4\cr
&Ours     &1.5 &\textbf{1.5} &\textbf{1.5} &\textbf{1.6}\cr
\midrule
\multirow{3}[2]{*}{\textbf {DialoGPT}}
&MLE      &1.3 &1.1 &0.7 &0.7\cr
&Standard &\textbf{1.5} &1.4 &1.2 &1.2\cr
&Ours     &1.4 &\textbf{1.5} &\textbf{1.6} &\textbf{1.6}\cr
\midrule
\multirow{3}[2]{*}{\textbf {T5}}
&MLE      &1.2 &0.9 &0.5 &0.6\cr
&Standard &1.1 &0.8 &0.6 &0.7\cr
&Ours     &\textbf{1.4} &\textbf{1.4} &\textbf{1.4} &\textbf{1.3}\cr
\midrule
\multirow{3}[2]{*}{\textbf {GODEL}}
&MLE      &1.5 &1.3 &0.8 &1.0\cr
&Standard &1.6 &1.2 &1.1 &1.1\cr
&Ours     &\textbf{1.7} &\textbf{1.6} &\textbf{1.7} &\textbf{1.6}\cr
\bottomrule
\end{tabular}
\caption{\label{t2}The final scores for each metric were calculated by taking the average of the annotator ratings. Each metric is measured per dialogue, and the best score in each metric is presented in bold. The Fleiss’ kappa \cite{fleiss1971measuring} score with human judges was approximately 0.29, which can be regarded as “fair agreement.”}
\end{table*}
\begin{figure*}[t]
\centering
\subfigure[CS]{
\includegraphics[width=7cm]{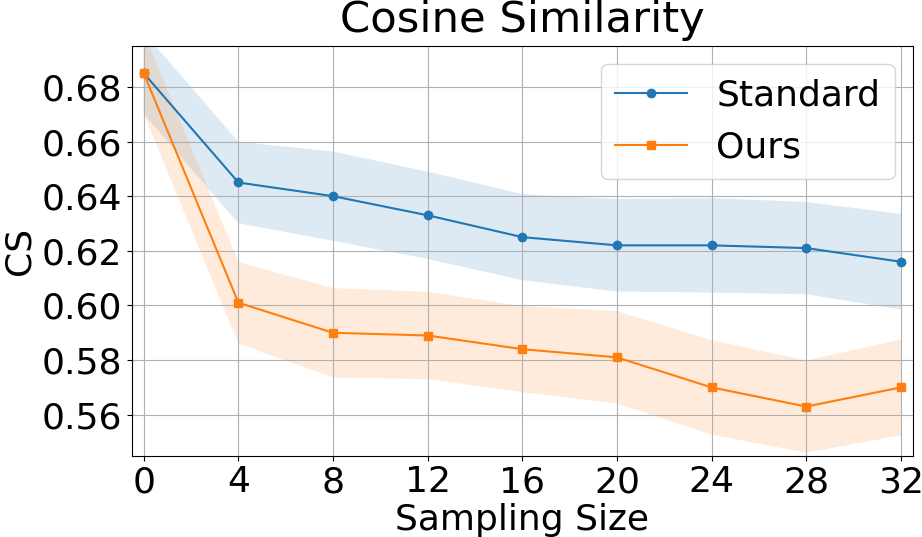}
}
\subfigure[SE]{
\includegraphics[width=7cm]{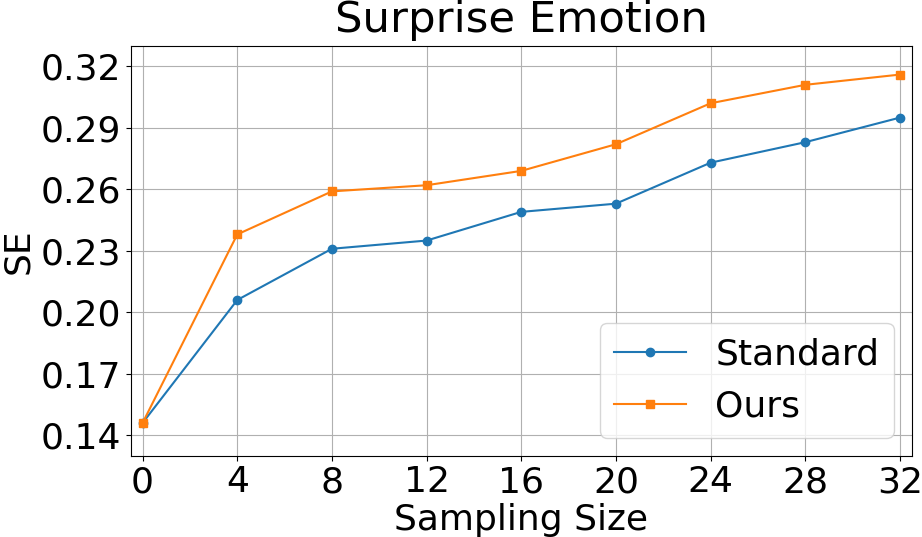}
}
\subfigure[LR]{
\includegraphics[width=7cm]{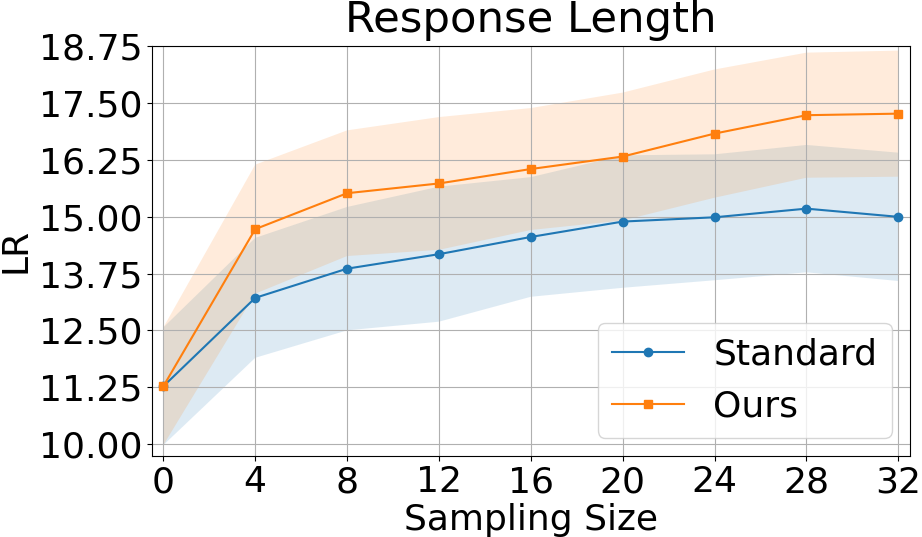}
}
\subfigure[AQ]{
\includegraphics[width=7cm]{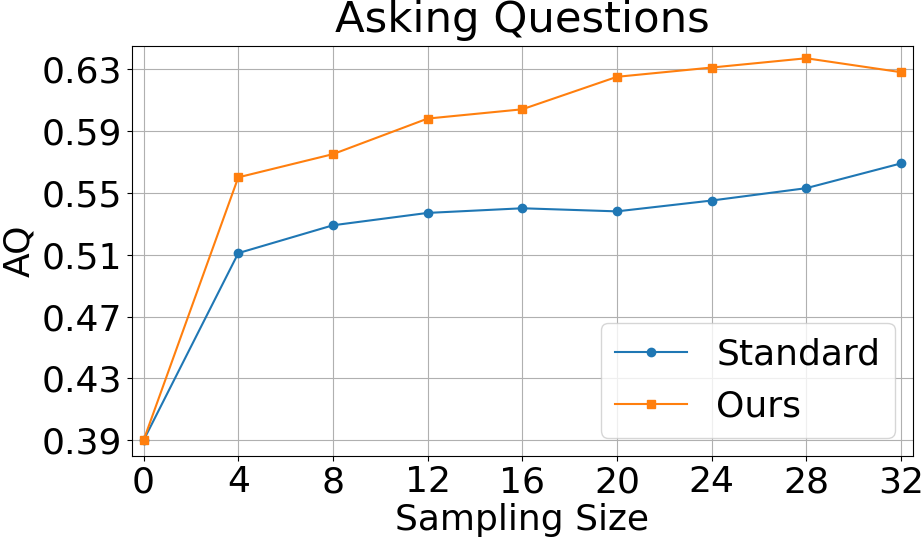}
}
\caption{\label{g1}The evolution of the agent's performance for each metric with the increased sampling size. The scale for the X-axis is a multiple of 4, and 0 represents the MLE without RL. Bands indicate half a standard deviation.}
\end{figure*}
\subsubsection{Automatic Metrics}
We apply the reward perspective designed in Section 2.3 to automatic evaluation. In particular, for the view of reward (2), we count the number of generated responses that contain a word that expresses surprise (i.e., Aha, Oh, Wow, Whoa, Gee, Really?, Amazing) by conservative string-matching heuristics. For the view of reward (4), we count the number of generated responses that contain a question word or a question mark. We denote CS, SE, LR, and AQ as cosine similarity, surprise emotion, response length, and asking questions, respectively. In our automatic evaluation and discussion, each metric is measured per response, and its score is obtained by taking the average across all samples.

\subsubsection{Human Metrics}
Ten native speakers were recruited to evaluate all agents trained using different methods. We asked the annotators to engage in a conversation with all agents regarding daily life topics (e.g., hobbies and interests) for at least five consecutive turns and rate their overall experience based on the following criteria. The scale of these metrics is [0, 1, 2].\\
\textbf{Quality} measures the coherence and grammatical accuracy of the agents’ responses. \emph{Score 0}: Most responses are incoherent or contain grammatical errors, thereby preventing the dialogue from proceeding. \emph{Score 1}: Although certain responses are incoherent or contain grammatical errors, the dialogue can continue. \emph{Score 2}: Only a few (or no) incoherent or grammatical errors in the responses, and the overall dialogue flows fluently.\\
\textbf{Informativeness} measures the diversity and hallucination of the agents’ responses. \emph{Score 0}: Most responses simply repeat information from the context or are generic. \emph{Score 1}: The information conflicts with common sense or contradicts the previous statement. \emph{Score 2}: Most responses have the appropriate information.\\
\textbf{Empathy} measures the degree to which agents respond with concern or affectivity. \emph{Score 0}: Most responses were short or showed little concern for the users in the dialogue. \emph{Score 1}: Although not very coherent, certain responses convey an emotional tone or ask a question. \emph{Score 2}: Certain responses are both coherent and show care for or emotional attachment to the user.\\
\textbf{Engagingness} measures the desire to engage the agents in a long conversation. \emph{Score 0}: The replies are lackluster, thereby making it difficult to sustain the dialogue. \emph{Score 1}: The responses are not particularly engaging, but they are fair for continuing the dialogue. \emph{Score 2}: The responses are engaging and have the potential to develop into a dialogue.

\subsubsection{Results}
We assessed the performance of our approach using dataset-based evaluation and compared it with baseline methods, which include a standard offline RL algorithm (i.e., Eq.(\ref{3}) and Eq.(\ref{2}), where Eq.(\ref{3}) is equivalent to our fine-grained Q-function, and it is referred to as the standard method in the following account) and MLE without RL. We also conducted a simulator-based evaluation by interacting with the user simulator Blenderbot \cite{roller2021recipes} to assess the performance of different methods in a long-term dialogue. Table \ref{t1} reports the automatic evaluation results. The standard method shows better performance than MLE, which can be credited to the policy improvement introduced by RL. In contrast, our approach achieved substantial gains in all metrics, thereby demonstrating the effectiveness of the dual-granularity Q-function. 

Table \ref{t2} summarizes the results of human evaluation. Despite the varied strengths and weaknesses of each agent according to individual human ratings, our approach exhibited markedly better results compared to baseline methods. Furthermore, RL-based agents displayed better proficiency than MLE-based agents in empathy and engagement by utilizing knowledge of the rewards outlined in Section 2.3. In terms of quality, most agents scored higher than other metrics because of the capacity of large-scale models to generate responses that are similar to human language and the stable one-step policy improvement of offline RL, which prevents the divergence of responses from human language. With regard to informativeness, upon analyzing instances of failure, we identified several agents that provided unrealistic information. Nevertheless, our approach generated more diverse responses, thereby resulting in a more favorable outcome than other methods. The interface and platform details for human evaluation are presented in Appendix \ref{app_C}.

\section{Discussion}
\subsection{Case Study}
To conduct a comprehensive qualitative comparison between our approach and the baseline methods, we randomly selected four dialogues with varying topics from the testing set and shortened each dialogue into four consecutive utterances, thereby yielding four contexts per dialogue. We then instructed all agents to generate a response for each context based on their respective training methods. The resulting cases are provided in Appendix \ref{app_D}. We found that our algorithm and the standard method generated longer responses compared to MLE, thereby indicating that RL-trained agents have better conversational engagement. Both the standard method and ours tend to ask questions, but our algorithm produces a more expressive tone of voice that conveys surprise, such as "Oh!," "Wow!," "Really?" and so on. Although our approach and the standard method scored similarly in the quality evaluation, upon examining these generated instances, it becomes evident that the responses generated by ours are slightly more coherent than those of the standard method. We consider that this may be attributed to the coarse-grained Q-function, which tends to determine the category of context-related actions, as we will explain later.

\subsection{Further Verification}
First, to validate Theorem 1 and illustrate the impact of sampling size on policy performance, we conducted an additional trial. Figure \ref{g1} presents a comparison of the performance between the GPT-2 agents trained using the standard method and our approach, which was under the simulator-based setting with varying numbers of response candidates. As suggested by our theoretical derivation, increasing the sample size generally led to better performance. Our algorithm outperformed the standard method even when using the same number of actions for policy improvement, thereby indicating its efficiency in iterating policy. This result emphasizes the significance of sample size as a constraint on policy performance and highlights the efficacy of our approach in addressing this issue.

Next, we sought to verify whether our approach satisfies the hypothesis $\mathbb{E}_{a\backsim\pi_\alpha(a|s)}[Q^{\pi}(s,a)]\ge\mathbb{E}_{a\backsim\pi_\beta(a|s)}[Q^{\pi}(s,a)]$ in Theorem 2. With this intention, our control generator, which relies on the coarse-grained Q-function to provide the optimal category, can be represented as $\pi_\alpha$, the agent that learned by the standard method can be considered as $\pi_\beta$, and $Q^{\pi}$ is the fine-grained Q-function. The expected value was approximated by averaging the Q-value of each sample. The results reveal that for the GPT-2, DialoGPT, T5, and GODEL agents, $\mathbb{E}_{a\backsim\pi_\alpha(a|s)}[Q^{\pi}(s,a)]$ increases by 8.76\%, 8.71\%, 10.14\%, and 9.61\%, respectively, compared to $\mathbb{E}_{a\backsim\pi_\beta(a|s)}[Q^{\pi}(s,a)]$. This indicates that the categories selected by the coarse-grained Q-function can produce responses with a higher Q-value, thereby supporting the hypothesis in Theorem 2. Overall, these findings emphasize the potential of employing a coarse-to-fine-grained approach to narrow the scope of action for policy improvement to enhance dialogue agent performance.

\begin{figure*}[t]
\centering
    \includegraphics[width=14.6cm]{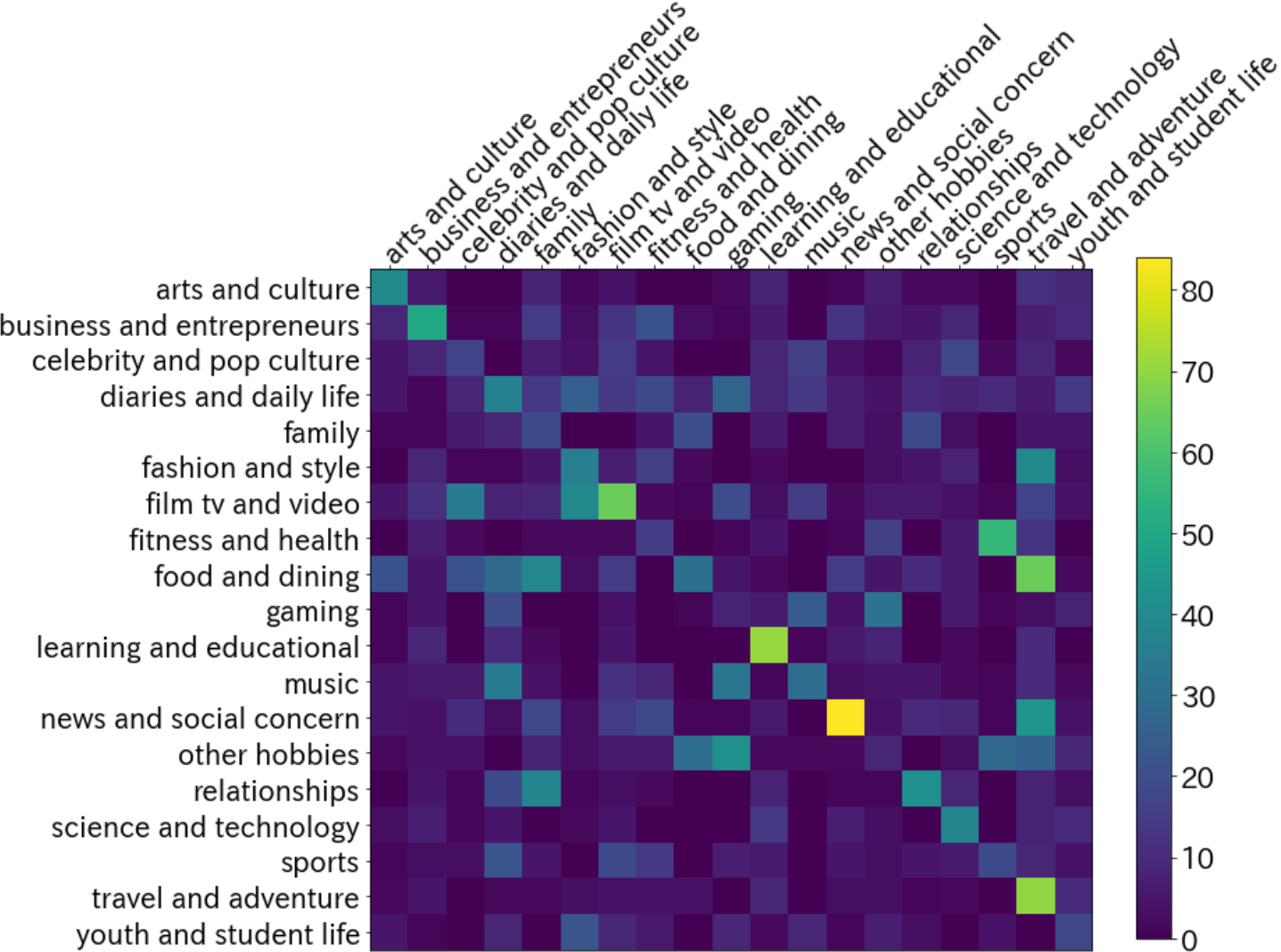}
    \caption{\label{g2} The label on the Y-axis represents the topic of each context, whereas the label on the X-axis represents the selected action category.}
\end{figure*}

Then, we examined the behavior of the coarse-grained Q-function in selecting action categories. We extracted 3000 contexts from the testing set and obtained the corresponding optimal action category by $\bar{a}^\ast=\mathop{\rm{arg\,max}}_{\bar{a}}Q_{\phi}(s,\bar{a})\quad\bar{a}\in\{\bar{a}_i\}^{N=19}_{i=1}$. The classifier was used to assign topics to these contexts. We then tallied the number of action categories selected by the coarse-grained Q-function for each context under each topic and this is visualized in Figure \ref{g2}. It was revealed that the selections made by the coarse-grained Q-function largely align with human intuition. For example, when the context pertains to the “film tv and video topic,” the coarse-grained Q-function often selects categories related to “fashion and style” and “celebrity and pop culture.” Similarly, when the context relates to the “other hobbies” topic, it tends to select categories related to “travel and adventure,” “gaming,” “music,” and “sports.” We also observed a significant proportion of choices concentrated along the matrix's main diagonal, thereby indicating its propensity to select action categories similar to the context topic. It is worth noting that "travel and adventure" also constitute a considerable part of the selected categories. After closely analyzing the dialogues in the corpus, we observed that discussions on travel are typically lengthier and require higher participation from both parties. This may have led the coarse-grained function to learn the extensive relevance of this category in dialogues.

Finally, we wanted to check if our control generator can generate responses for the specified category. For each $\bar{a}\in\{\bar{a}_i\}^{N=19}_{i=1}$, we used $\pi_\psi(a|s,\bar{a})$ to obtain a response and then used the classifier to determine if this response belongs to $\bar{a}$. The results reveal that the percentage of accurate responses to a given category for the control generator performed by GPT-2, DialoGPT, T5, and GODEL are 28.16\%, 29.89\%, 35.42\%, and 36.37\%, respectively. This ratio is significantly higher than the correct ratio obtained by randomly generating responses over 19 categories (i,e., 1/19 $\approx$ 5.26\%).

\section{Related Work}
Since the RL for dialogue requires multiple steps of expensive human interaction, several prior studies have updated the agent's policy by the self-play method or the interaction with simulators \cite{li2016deep,shah2018bootstrapping,peng2018deep,liu2020you}. However, these online RL methods suffer from the issue of diverging from human language \cite{lewis2017deal,zhao2019rethinking,jang2020bayes}. On the other hand, offline RL \cite{fujimoto2019off,kumar2020conservative,brandfonbrener2021offline,kostrikovoffline} eliminates all need for environmental interaction or user simulators, instead of operating purely on static datasets of prior human interaction. There are many closely related works \cite{jaques2019way,jaques2020human,snell2022offline,cohen2022dynamic,verma2022chai,jang2022gpt} based on offline RL that lead to policy improvement via behavior cloning of self-generated utterances, which inherits the ability of pre-trained language models to generate human-like responses. In RL parlance, such methods could be regarded as policy extraction with approximate dynamic programming. Nevertheless, unlike RL tasks in which the actions are finite, such as Atari games \cite{mnih2015human}, the dialogue setting is hard to explore all probability space. Therefore, the policy obtained through the aforementioned methods is suboptimal. Our dual-granularity Q-function focuses on the more structured action choices to effectively implement policy improvement.

\section{Conclusion and Future Research}
This paper presented a dual-granularity Q-function for mitigating suboptimal policy improvement due to the hard-to-traverse action space in RL, and we applied our method to the dialogue generation task. Theoretical and experimental results demonstrate the reliability of our algorithm, which significantly enhances the performance of the dialogue agent. Moving forward, we intend to design additional abstract categories for actions, such as those based on sentence embedding, to allow the coarse-grained Q-function to account for not only content but also utterance structure and expression. We will also investigate the affinity between the number of action categories and policy improvement. Ultimately, we plan to apply our algorithm to other online RL tasks in NLP to verify its universal applicability.

\bibliography{anthology,custom}
\bibliographystyle{acl_natbib}

\appendix
\section{Appendix}
\label{app_A}
\subsection*{Proofs of Theorem}
See Preliminary Derivation, Proof of Theorem 1, and Proof of Theorem 2 on the next page.
\begin{figure*}
\begin{equation}
\begin{split}
&{\rm \textbf{Preliminary Derivation}}:\\
&\mathbb{E}_{\pi^{\prime}}[\mathop{{\rm max}}\limits_{a\in\{a_i\}^L_{i=1}}Q^{\pi}(s,a)]\\
&=\mathbb{E}_{\pi^{\prime}}[R(s_t,\mathop{\rm{arg\,max}}\limits_{a\in\{a_i\}^L_{i=1}}Q^{\pi}(s_t,a))+\gamma R(s_{t+1},a_{t+1})+\gamma^2 R(s_{t+2},a_{t+2})+...|s_t=s,\{a_n\backsim\pi(a|s_n)\}_{n=t+1}^T]\\
&=\mathbb{E}_{\pi^{\prime}}[R(s_t,\mathop{\rm{arg\,max}}\limits_{a\in\{a_i\}^L_{i=1}}Q^{\pi}(s_t,a))+\gamma V^{\pi}(s_{t+1})|s_t=s,a\backsim\pi^{\prime}(a|s_t)]\\\\
&\textbf{Proof\;of\;Theorem 1.}\\
&{\rm Premise}:\quad\forall s, \,\pi^{\prime}_L(\cdot|s)=\mathop{\rm{arg\,max}}\limits_{a\in \{a_i\}^L_{i=1}\backsim\pi(a|s)}Q^{\pi}(s,a)\\
&{\rm Lemma}\,:\quad{\rm if}\;N\ge M\ge 1,\;{\rm then}\;\mathbb{E}_{\pi}[\mathop{{\rm max}}\limits_{a\in\{a_i\}^N_{i=1}}Q^{\pi}(s,a)]\ge\mathbb{E}_{\pi}[\mathop{{\rm max}}\limits_{a\in\{a_i\}^M_{i=1}}Q^{\pi}(s,a)]\\
&V^{\pi^{\prime}_N}(s)\\
&=\mathbb{E}_{\pi^{\prime}_N}[R(s_t,a_{t}\backsim\pi^{\prime}_N(a|s_t))+\gamma R(s_{t+1},a_{t+1}\backsim\pi^{\prime}_N(a|s_{t+1}))+...|s_t=s]\\
&=\mathbb{E}_{\pi}[R(s_t,\mathop{\rm{arg\,max}}\limits_{a\in\{a_i\}^N_{i=1}}Q^{\pi}(s_t,a))+\gamma R(s_{t+1},\mathop{\rm{arg\,max}}\limits_{a\in\{a_i\}^N_{i=1}}Q^{\pi}(s_{t+1},a))+...|s_t=s]\\
&=\mathbb{E}_{\pi}[\mathop{{\rm max}}\limits_{a\in\{a_i\}^N_{i=1}}Q^{\pi}(s,a)-\gamma V^{\pi}(s_{t+1})|s_t=s]+\gamma\mathbb{E}_{\pi}[\mathop{{\rm max}}\limits_{a\in\{a_i\}^N_{i=1}}Q^{\pi}(s_{t+1},a)-\gamma V^{\pi}(s_{t+2})|s_t=s]+...\\
&\ge\mathbb{E}_{\pi}[\mathop{{\rm max}}\limits_{a\in\{a_i\}^M_{i=1}}Q^{\pi}(s,a)-\gamma V^{\pi}(s_{t+1})|s_t=s]+\gamma\mathbb{E}_{\pi}[\mathop{{\rm max}}\limits_{a\in\{a_i\}^M_{i=1}}Q^{\pi}(s_{t+1},a)-\gamma V^{\pi}(s_{t+2})|s_t=s]+...\\
&=\mathbb{E}_{\pi}[R(s_t,\mathop{\rm{arg\,max}}\limits_{a\in\{a_i\}^M_{i=1}}Q^{\pi}(s_t,a))+\gamma R(s_{t+1},\mathop{\rm{arg\,max}}\limits_{a\in\{a_i\}^M_{i=1}}Q^{\pi}(s_{t+1},a))+...|s_t=s]\\
&=\mathbb{E}_{\pi^{\prime}_M}[R(s_t,a_{t}\backsim\pi^{\prime}_M(a|s_t))+\gamma R(s_{t+1},a_{t+1}\backsim\pi^{\prime}_M(a|s_{t+1}))+...|s_t=s]=V^{\pi^{\prime}_M}(s)\\\\
&\textbf{Proof\;of\;Theorem 2.}\\
&{\rm Premise\,1}:\quad\forall s, \,\pi^{\prime}_1(\cdot|s)=\mathop{\rm{arg\,max}}\limits_{a\in \{a_i\}^L_{i=1}\backsim\pi_\alpha(a|s)}Q^{\pi}(s,a),\;\;\pi^{\prime}_2(\cdot|s)=\mathop{\rm{arg\,max}}\limits_{a\in \{a_i\}^L_{i=1}\backsim\pi_\beta(a|s)}Q^{\pi}(s,a)\\
&{\rm Premise\,2}:\quad\mathbb{E}_{a\backsim\pi_\alpha(a|s)}[Q^{\pi}(s,a)]\ge\mathbb{E}_{a\backsim\pi_\beta(a|s)}[Q^{\pi}(s,a)],\;\;\sigma^2_{a\backsim\pi_\alpha(a|s)}[Q^{\pi}(s,a)]\approx\sigma^2_{a\backsim\pi_\beta(a|s)}[Q^{\pi}(s,a)]\\
&{\rm Lemma}\;\;\;:\;\quad\because\;{\rm Premise\,2}\;\;\therefore\;\mathbb{E}_{\pi_\alpha}[\mathop{{\rm max}}\limits_{a\in\{a_i\}^L_{i=1}}Q^{\pi}(s,a)]\ge\mathbb{E}_{\pi_\beta}[\mathop{{\rm max}}\limits_{a\in\{a_i\}^L_{i=1}}Q^{\pi}(s,a)]\\
&V^{\pi^{\prime}_1}(s)\\
&=\mathbb{E}_{\pi^{\prime}_1}[R(s_t,a_{t}\backsim\pi^{\prime}_1(a|s_t))+\gamma R(s_{t+1},a_{t+1}\backsim\pi^{\prime}_1(a|s_{t+1}))+...|s_t=s]\\
&=\mathbb{E}_{\pi_\alpha}[R(s_t,\mathop{\rm{arg\,max}}\limits_{a\in\{a_i\}^L_{i=1}}Q^{\pi}(s_t,a))+\gamma R(s_{t+1},\mathop{\rm{arg\,max}}\limits_{a\in\{a_i\}^L_{i=1}}Q^{\pi}(s_{t+1},a))+...|s_t=s]\\
&=\mathbb{E}_{\pi_\alpha}[\mathop{{\rm max}}\limits_{a\in\{a_i\}^L_{i=1}}Q^{\pi}(s,a)-\gamma V^{\pi}(s_{t+1})|s_t=s]+\gamma\mathbb{E}_{\pi_\alpha}[\mathop{{\rm max}}\limits_{a\in\{a_i\}^L_{i=1}}Q^{\pi}(s_{t+1},a)-\gamma V^{\pi}(s_{t+2})|s_t=s]+...\\
&\ge\mathbb{E}_{\pi_\beta}[\mathop{{\rm max}}\limits_{a\in\{a_i\}^L_{i=1}}Q^{\pi}(s,a)-\gamma V^{\pi}(s_{t+1})|s_t=s]+\gamma\mathbb{E}_{\pi_\beta}[\mathop{{\rm max}}\limits_{a\in\{a_i\}^L_{i=1}}Q^{\pi}(s_{t+1},a)-\gamma V^{\pi}(s_{t+2})|s_t=s]+...\\
&=\mathbb{E}_{\pi_\beta}[R(s_t,\mathop{\rm{arg\,max}}\limits_{a\in\{a_i\}^L_{i=1}}Q^{\pi}(s_t,a))+\gamma R(s_{t+1},\mathop{\rm{arg\,max}}\limits_{a\in\{a_i\}^L_{i=1}}Q^{\pi}(s_{t+1},a))+...|s_t=s]\\
&=\mathbb{E}_{\pi^{\prime}_2}[R(s_t,a_{t}\backsim\pi^{\prime}_2(a|s_t))+\gamma R(s_{t+1},a_{t+1}\backsim\pi^{\prime}_2(a|s_{t+1}))+...|s_t=s]=V^{\pi^{\prime}_2}(s)
\end{split}
\nonumber
\end{equation}
\end{figure*}

\section{Appendix}
\label{app_B}
\subsection*{Details regarding Rewards}
(1) Cosine Similarity: We manually created a list of dull responses consisting of utterances such as “I don’t know,” etc., which are short and frequently occur in generation models. We penalize the cosine similarity between the agent’s response and the dull responses to avoid the generation of dull responses by the agent. This score is computed by leveraging a state-of-the-art sentence embedding model \cite{conneau2017supervised}, and the score range is 0 to 1. Although there are more ways to generate a dull response, similar expressions are likely to fall into an adjacent vector space. The user “keeps away” from the utterances in the list, thereby also keeping away from other similar dull responses.\\
(2) Surprise Emotion: Since each utterance in the DailyDialog dataset was annotated by one of six universal emotions in human beings, we used the emotional label of “Surprise” to allocate a reward to each sentence, for which the scale is [0, 1].\\
(3)	Response Length: From an empathetic standpoint, we prefer that the agents generate responses that are more elaborate and longer. The reward is defined in the following manner.\par\quad
\textbf{if} \quad the number of generated tokens < 5:\par\qquad\quad
reward = -0.2\par\quad
\textbf{elif} \;the number of generated tokens < 10:\par\qquad\quad
  reward = 0\par\quad
\textbf{elif} \;the number of generated tokens < 15:\par\qquad\quad
reward = 0.2\par\quad
\textbf{else}:\par\qquad\quad
reward = 0.5\\
(4)	Asking Questions: Each utterance in the DailyDialog dataset was also labeled as one of four dialogue act classes. We used the act label of “Questions” to allocate a reward to each sentence. The scale of this reward is [0, 1].

\section{Appendix}
\label{app_C}
\subsection*{Details regarding Interactive}
\begin{figure*}[t]
\centering
\subfigure[Chat Interface 1]{
\includegraphics[width=5cm]{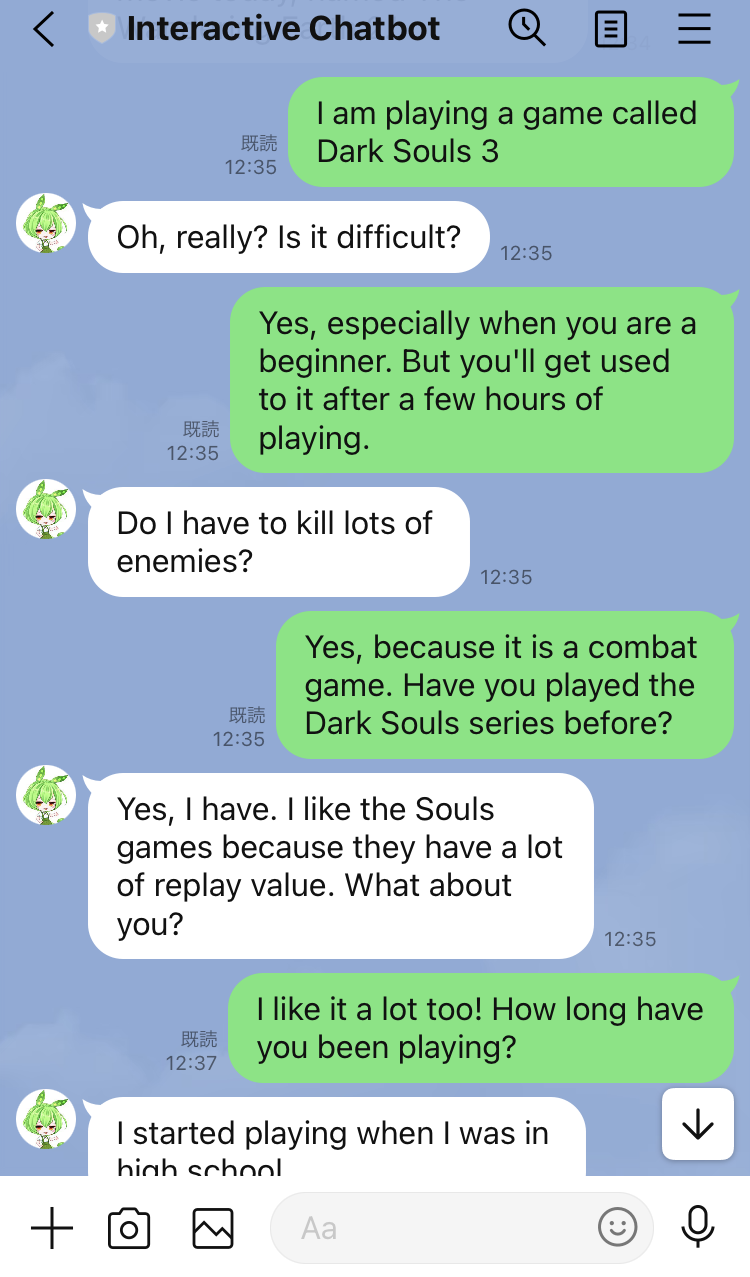}
}
\subfigure[Chat Interface 2]{
\includegraphics[width=5cm]{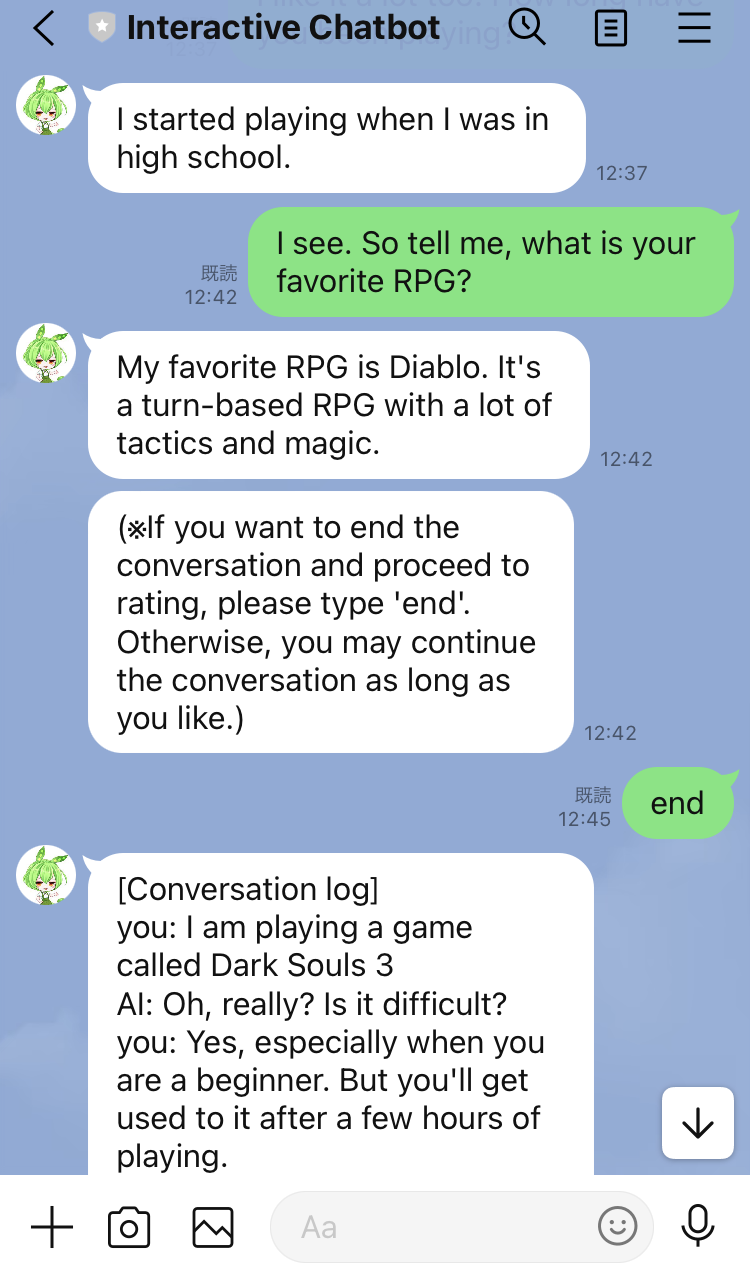}
}
\subfigure[Ratings Interface 1]{
\includegraphics[width=5cm]{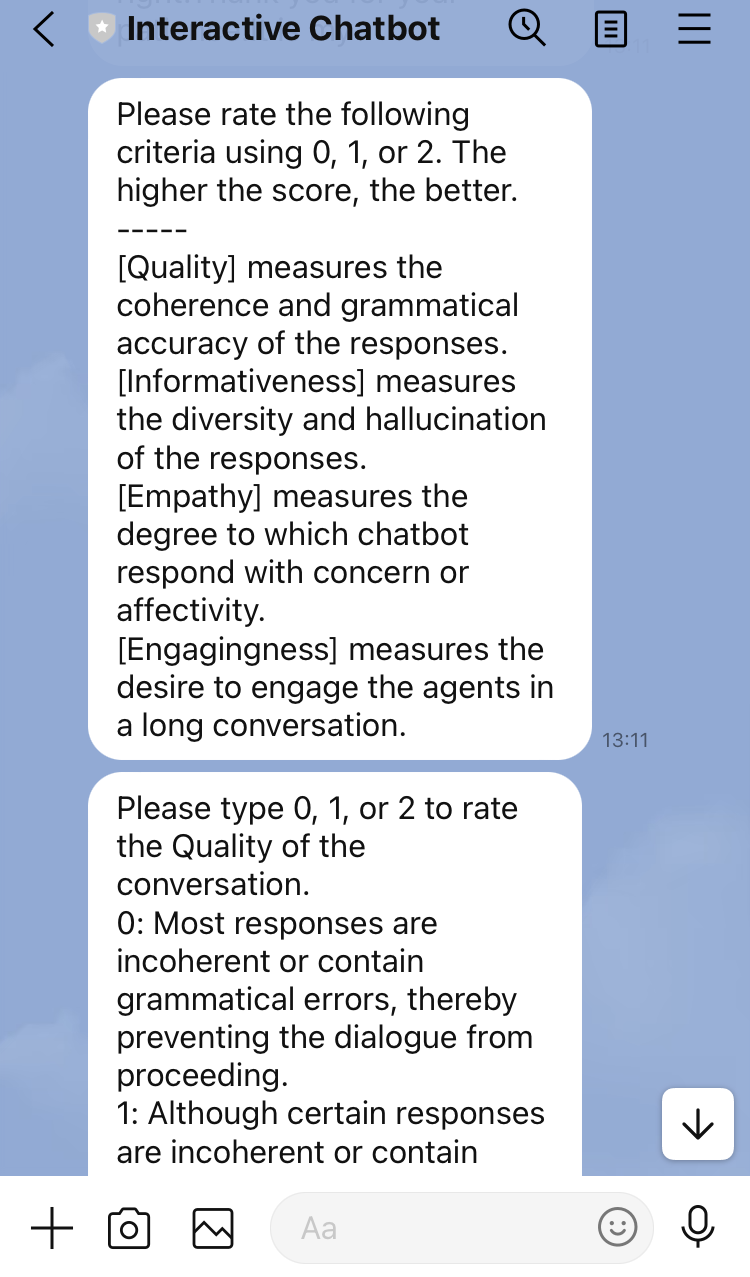}
}
\subfigure[Ratings Interface 2]{
\includegraphics[width=5cm]{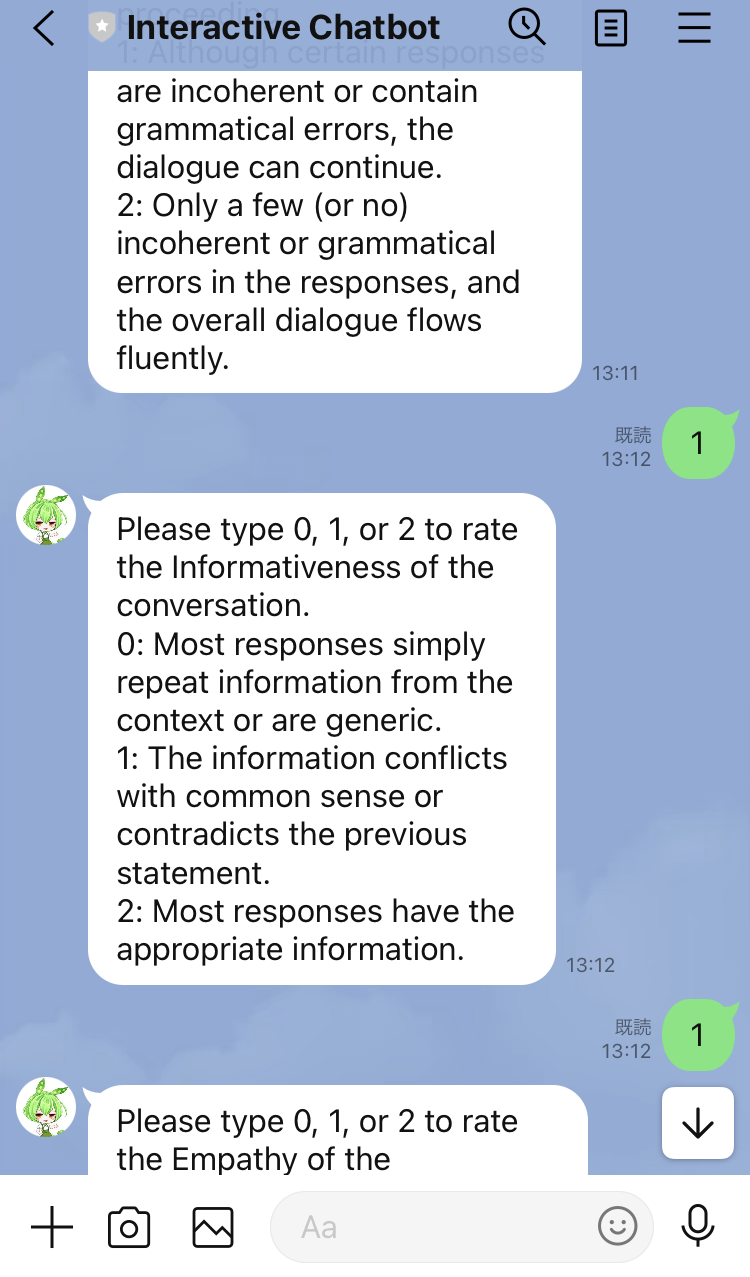}
}
\subfigure[Ratings Interface 3]{
\includegraphics[width=5cm]{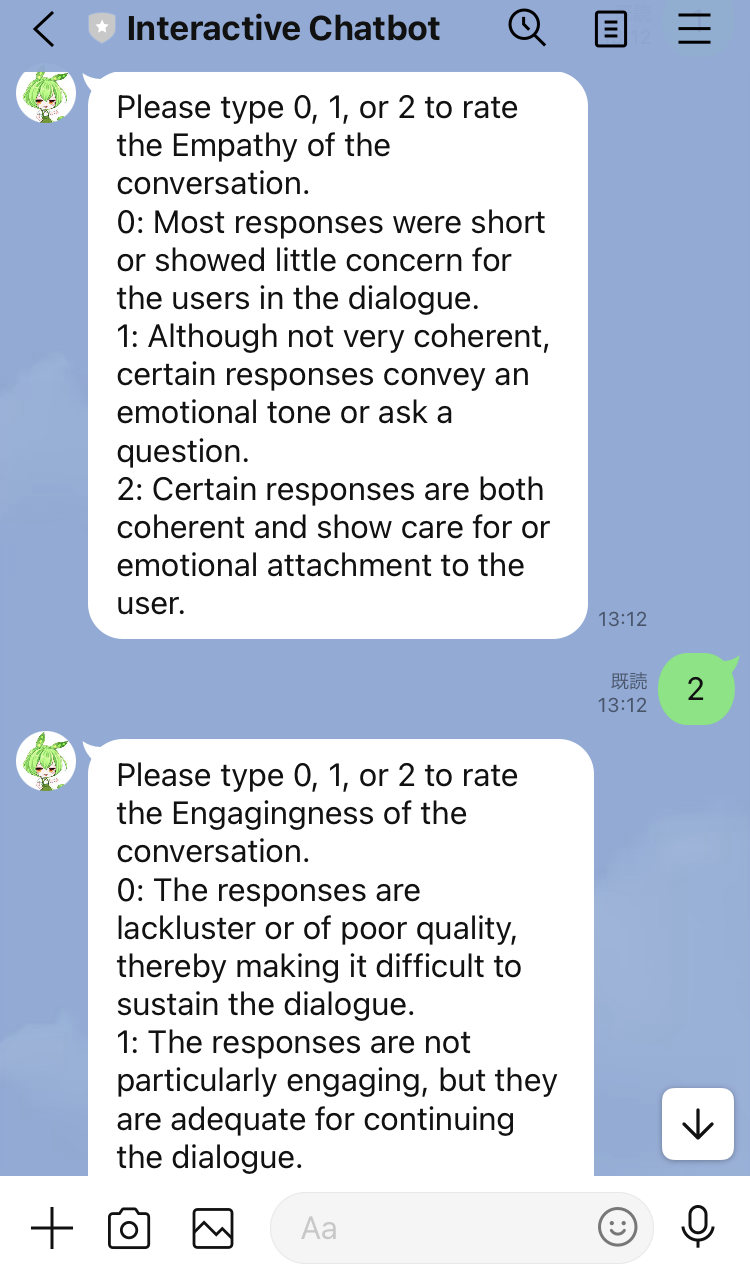}
}
\subfigure[Ratings Interface 4]{
\includegraphics[width=5cm]{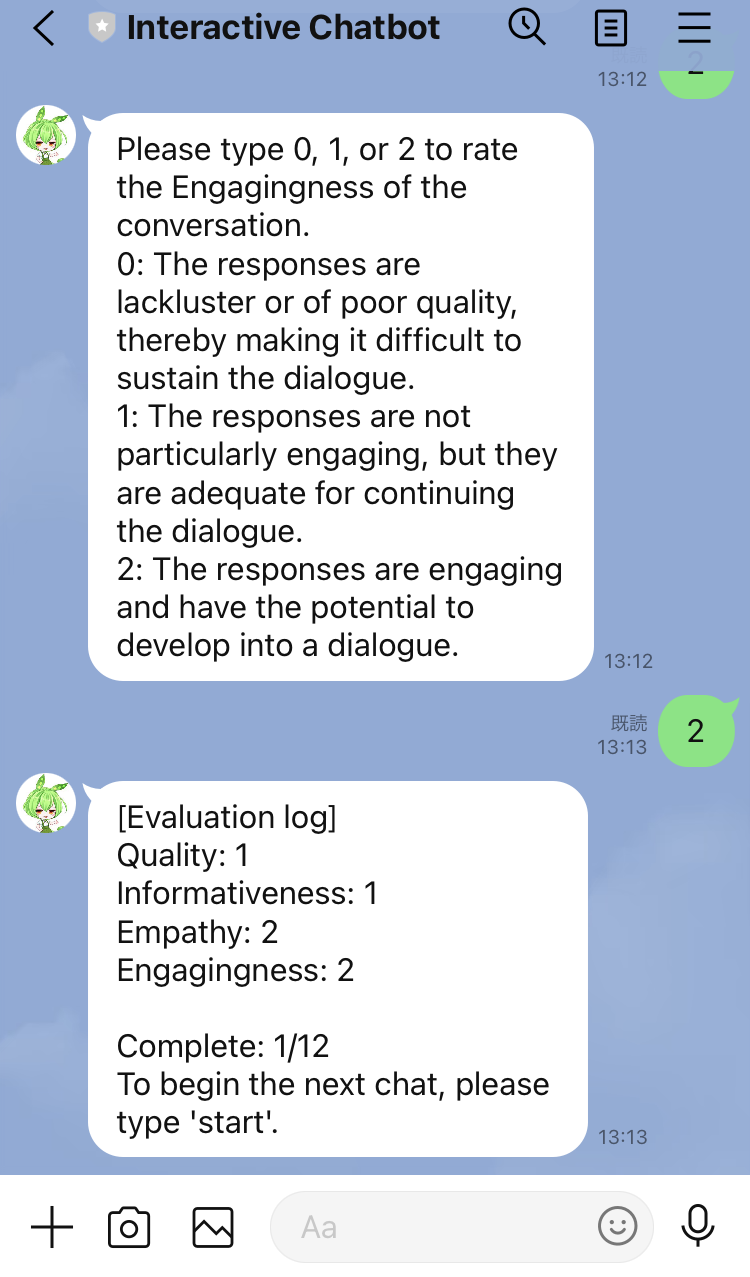}
}
\caption{\label{g3}The chat and rating interfaces used for human evaluation.}
\end{figure*}

We utilized LINE\footnote{https://line.me/en/} as our platform and set up a specialized account for human evaluation, which evaluators accessed through their smart device to interact with each agent and provide ratings. The server was equipped with an NVIDIA A6000 (48G) graphics card, and the program was developed using the LINE Messaging API SDK and ran continuously in the background, ready to receive requests at any time. Each agent extended an abstract class that defined key methods for conversation generation and was registered to a dictionary via a decorator. To ensure a randomized order of appearance of agents for annotators during the evaluation process, we implemented a randomized selection of dictionary indices. Furthermore, due to the substantial startup times of the agents, all agents were kept in memory at all times in the background process. The current configuration was able to support hundreds of simultaneous users and concurrently host more than 20 agents. Figures \ref{g3} (a) and (b) depict the conversation interface utilized by annotators for interacting with the agents during the human evaluation process. This interface enabled participants to engage in a dialogue of at least five turns before initiating the rating phase by entering the word "end." In contrast, Figures \ref{g3} (c), (d), (e), and (f) exhibit the interface used for rating the agents after having a conversation of at least five turns with them as part of the human evaluation.

\section{Appendix}
\label{app_D}
\subsection*{Case Study}
Figure \ref{g4} presents the varied responses generated by different methods across four agents. The ground truth for responding to each user utterance is the next user utterance in the dialogue.
\begin{figure*}[t]
\centering
\subfigure[GPT-2]{
\includegraphics[width=15cm]{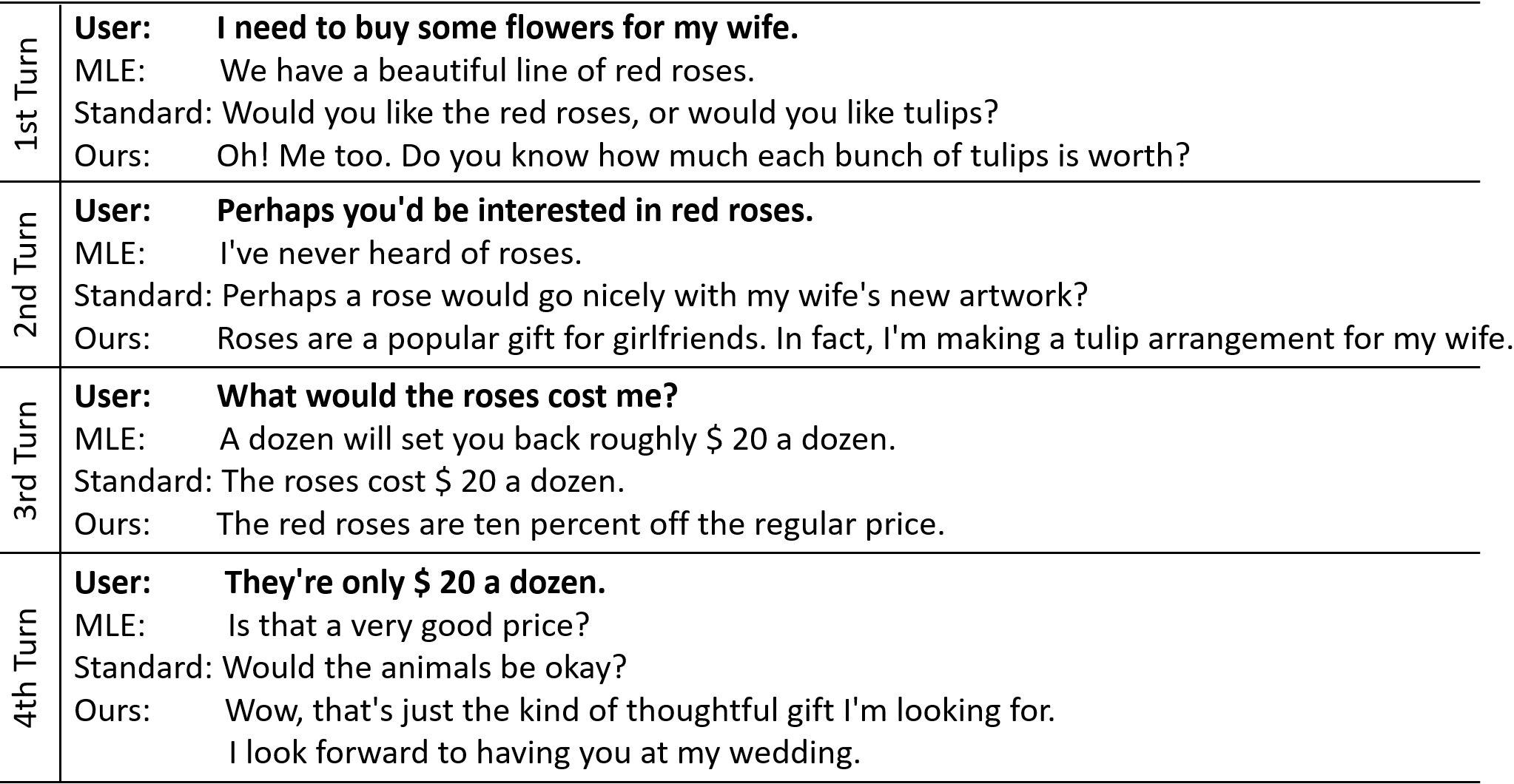}
}
\subfigure[DialoGPT]{
\includegraphics[width=15cm]{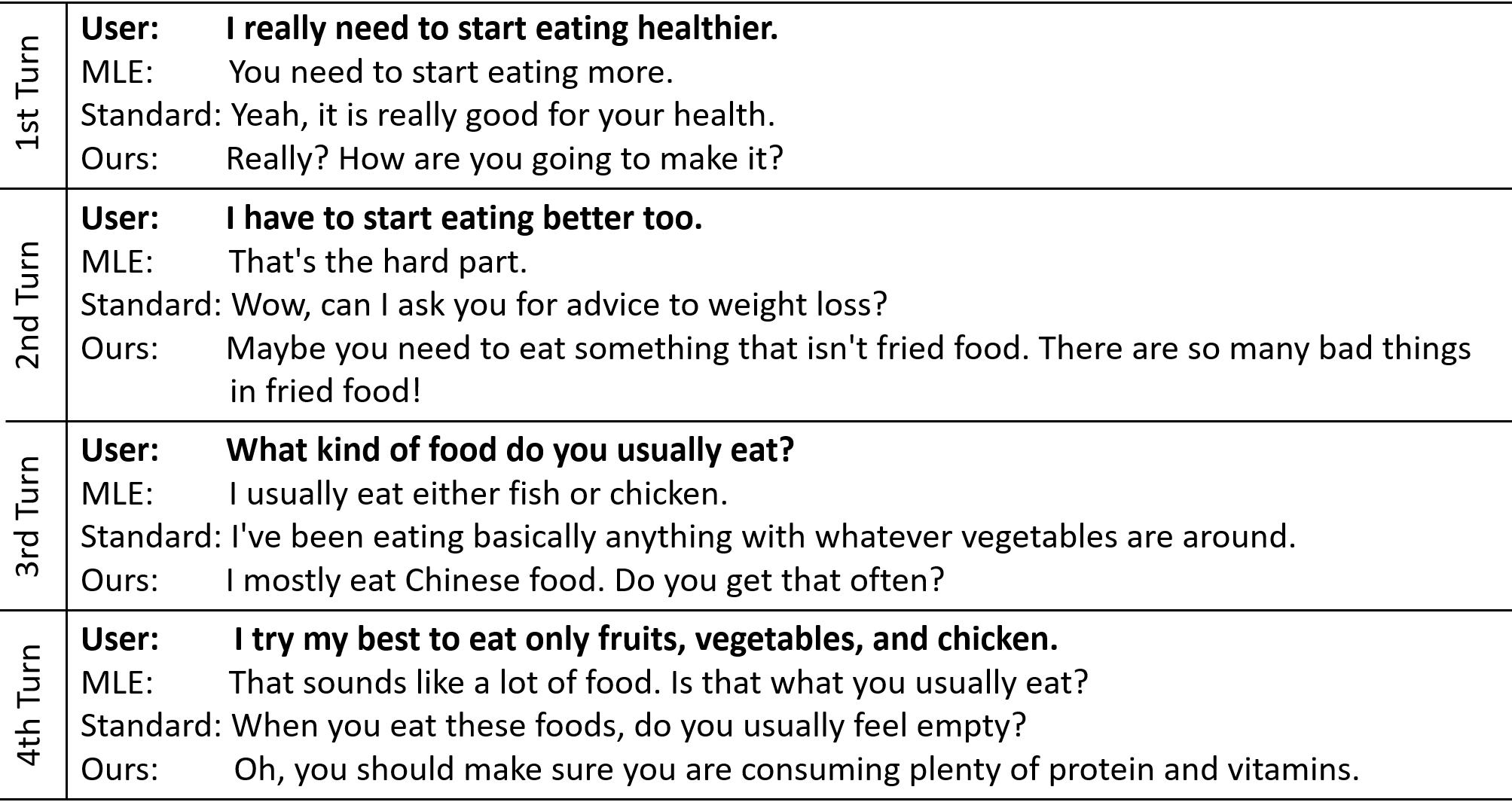}
}
\end{figure*}
\addtocounter{figure}{-1}
\begin{figure*} 
\addtocounter{figure}{1}
\centering
\subfigure[T5]{
\includegraphics[width=15cm]{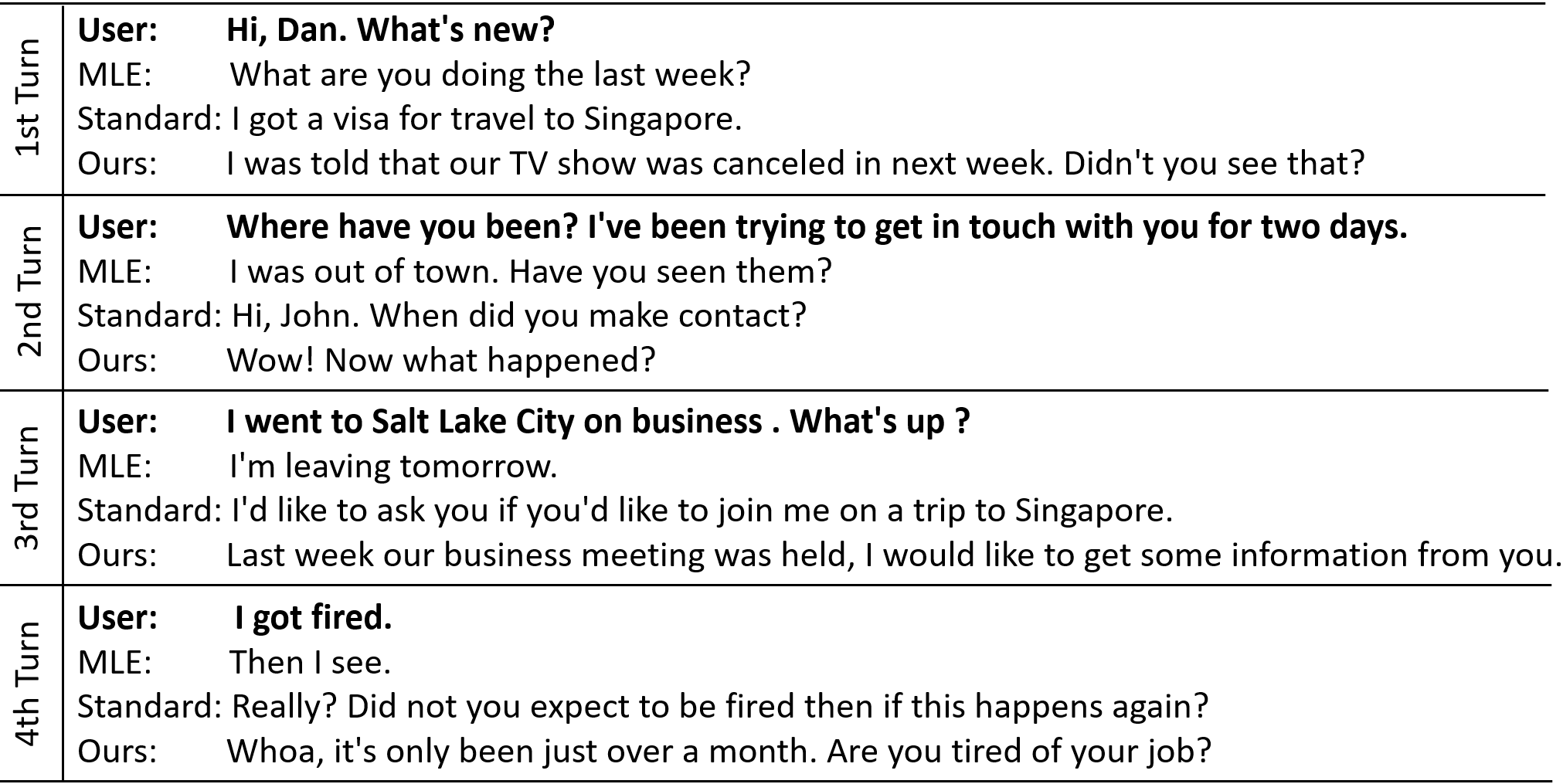}
}
\subfigure[GODEL]{
\includegraphics[width=15cm]{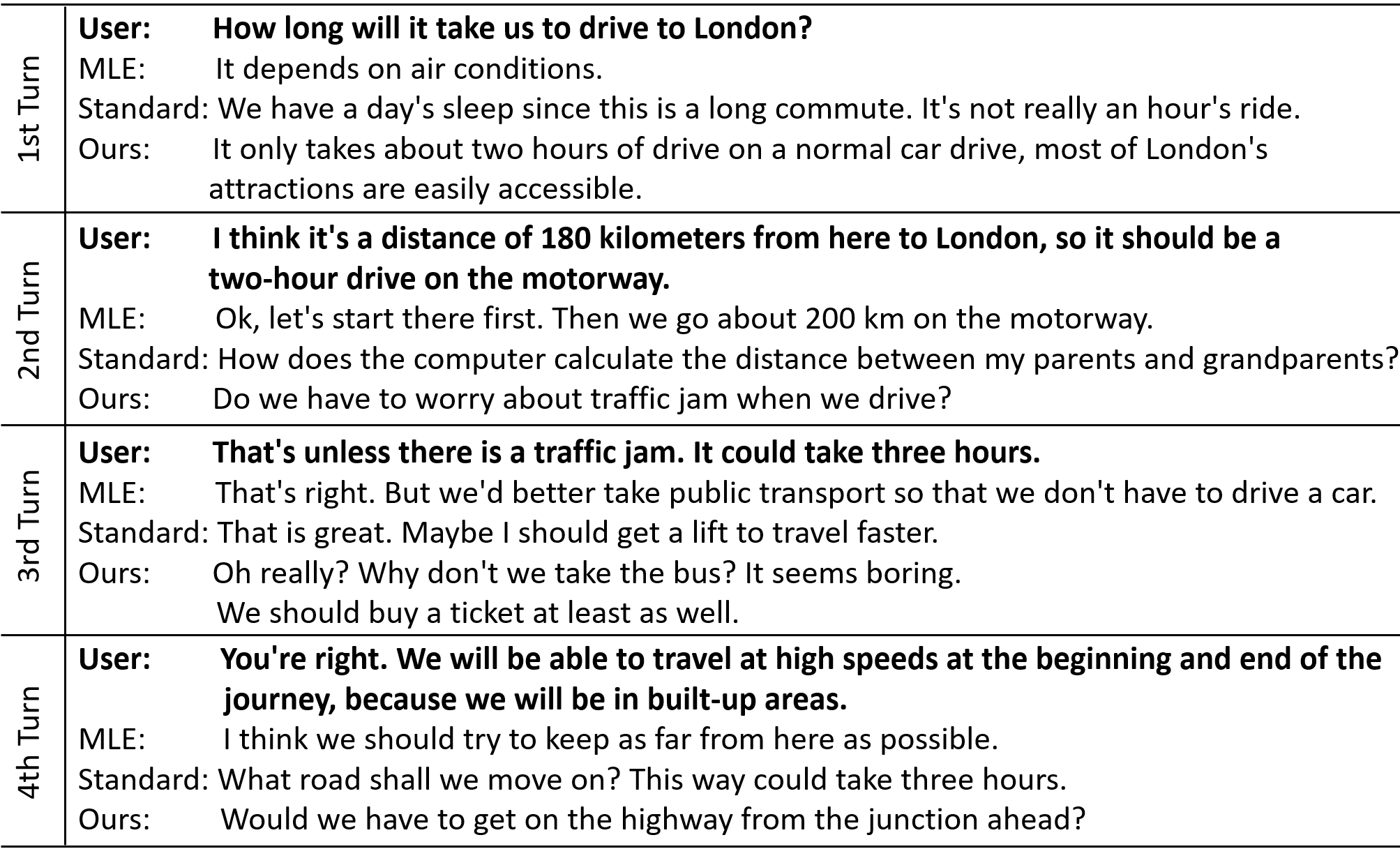}
}
\caption{\label{g4}Case Study.}
\end{figure*}

\end{document}